# GRU-D-Weibull: A Novel Real-Time Individualized Endpoint Prediction

Dr. Xiaoyang Ruan [1,2]; Dr. Liwei Wang [1,2]; Dr. Charat Thongprayoon [3]; Dr. Wisit Cheungpasitporn [3]; Dr. Hongfang Liu[1,2]*

[1] McWilliams School of Biomedical Informatics, University of Texas Health Science Center at Houston

[2] Department of Artificial Intelligence & Informatics, Mayo Clinic, Rochester, MN, United States

[3] Department of Internal Medicine, Mayo Clinic, Rochester, MN, United States

Corresponding Author:

Hongfang Liu, PhD

McWilliams School of Biomedical Informatics, University of Texas Health Science Center at Houston

7000 Fannin St, Houston, TX 77030

United States

Phone 1 507 316 2936

Email: hongfang.liu@uth.tmc.edu

**Abstract**

**Background:** Accurate risk prediction models for individual level endpoint (e.g., death), or time-to-endpoint are highly desirable in clinical practice. However, most of the existing predictive modeling strategies leverage well curated data sets with limited real-world implementation potential in practice.

**Methods:** We propose a novel predictive modeling approach, GRU-D-Weibull, which models Weibull distribution leveraging gated recurrent units with decay (GRU-D), for real-time individualized endpoint prediction and population level risk management using electronic health records (EHRs).

**Experiments:** We systematically evaluated the performance and showcased the clinical utility of the proposed approach through individual level endpoint prediction using a cohort of patients with chronic kidney disease stage 4 (CKD4). A total of 536 features including ICD/CPT codes, medications, lab tests, vital measurements, and demographics were retrieved for 6,879 CKD4 patients. The performance metrics including C-index, L1-loss, Parkes' error, and predicted survival probability at time of event were compared between GRU-D-Weibull and other alternative approaches including accelerated failure time model (AFT), XGBoost(AFT), random survival forest (RSF), and Nnet-survival. Both in-process and post-process calibrations were experimented on GRU-D-Weibull generated survival probabilities.

**Results:** GRU-D-Weibull demonstrated C-index of ~0.7 at index date, which increased to ~0.77 at 4.3 years of follow-up, comparable to that of RSF. GRU-D-Weibull achieved absolute L1-loss of ~1.1 years (sd≈0.95) at CKD4 index date, and a minimum of ~0.45 year (sd≈0.3) at 4 years of follow-up, comparing to second-ranked RSF of ~1.4 years (sd ≈1.1) at index date and ~0.64 years (sd≈0.26) at 4 years. Both significantly outperform competing approaches. GRU-D-Weibull constrained predicted survival probability at time of event to a remarkably smaller and more fixed range than competing models throughout follow-up. Significant correlations were observed between prediction error and missing proportions of all major categories of input features at index date (Corr ~0.1 to ~0.3), which faded away within 1 year after index date as more data became available. Through post training recalibration, we achieved a close alignment between the predicted and observed survival probabilities across multiple prediction horizons at different time points during follow-up.

**Conclusion:** Our proposed GRU-D-Weibull approach surpasses other popular alternative methods by effectively handling missing data and providing both probability and point estimates for diverse prediction horizons during follow-up. The experiment highlights the significant potential of GRU-D-Weibull as a next-generation architecture for endpoint risk management, utilizing real-time clinical data to generate various endpoint estimates for monitoring. Further studies involving clinicians are necessary to investigate the integration of this approach into clinical workflows and assess its impact on decision-making processes and patient outcomes.

**Abstract (shorter version)**
Accurate prediction models for individual-level endpoints and time-to-endpoints are crucial in clinical practice. In this study, we propose a novel approach, GRU-D-Weibull, which combines gated recurrent units with decay (GRU-D) to model the Weibull distribution. Our method enables real-time individualized endpoint prediction and population-level risk management. Using a cohort of 6,879 patients with stage 4 chronic kidney disease (CKD4), we evaluated the performance of GRU-D-Weibull in endpoint prediction. The C-index of GRU-D-Weibull was ~0.7 at the index date and increased to ~0.77 after 4.3 years of follow-up, similar to random survival forest. Our approach achieved an absolute L1-loss of ~1.1 years (SD≈0.95) at the CKD4 index date and a minimum of ~0.45 years (SD≈0.3) at 4 years of follow-up, outperforming competing methods significantly. GRU-D-Weibull consistently constrained the predicted survival probability at the time of an event within a smaller and more fixed range compared to other models throughout the follow-up period. We observed significant correlations between the error in point estimates and missing proportions of input features at the index date (correlations from ~0.1 to ~0.3), which diminished within 1 year as more data became available. By post-training recalibration, we successfully aligned the predicted and observed survival probabilities across multiple prediction horizons at different time points during follow-up. Our findings demonstrate the considerable potential of GRU-D-Weibull as the next-generation architecture for endpoint risk management, capable of generating various endpoint estimates for real-time monitoring using clinical data.

**Background**
Accurate prediction of the risk of reaching an endpoint and providing a point estimate of the time-to-endpoint at the individual level are of utmost importance for effective clinical decision making. This is particularly significant in the context of major public health issues such as chronic kidney disease (CKD), which affects a substantial portion of the global population (8% to 16%) [1–7] and has been associated with a rising mortality rate [8–10]. CKD is responsible for approximately 1.2 million deaths worldwide [11], underscoring the critical need for precise individual-level predictions in this clinical domain. However, individualized predictions based on current statistical or machine learning models are of very limited use due to data quality issues in clinical data, intrinsic large statistical variations, and changes of the trajectory of disease progression due to interventions. Take CKD as an example, existing endpoint predictive models including decision tree [12] for individualized above/below median survival time prediction of hemodialysis patients, COX proportional hazards model [13] based static risk scoring for progression of CKD to kidney failure, ensemble of feed forward multilayer perceptron [14] to evaluate ESRD risk as a binary outcome of IgA nephropathy patients, multi-layer perceptron [15] to predict CKD 5-year survival as a binary outcome, and others [16] [17] [18] [19] [20]. The current methodologies employed for prediction in clinical settings have notable limitations, such as reliance on static inputs, ability to only perform binary or multi-class classification, restricted prediction horizons, and inability to effectively handle missing data. Additionally, constructing and maintaining separate survival models at different disease progression time points is impractical due to the dynamic nature of diseases, the challenges associated with sample collection, and the inherent sparsity and missingness of data that are commonly encountered during extended follow-up periods.

In recent years, several deep learning architectures have emerged for survival prediction in the medical field. Some notable examples include DeepSurv [21], which employs a Cox proportional hazards deep neural network to offer personalized treatment recommendation. Cox-nnet [22] utilizes a similar architecture as DeepSurv but is specifically optimized for high throughput omics data. Nnet-survival [23], on the other hand, employs a deep feed-forward network to estimate discrete-time survival probabilities. One advantage of deep learning models is their ability to bypass the need for extensive feature engineering and their ability to capture complex interactions between features without prior medical knowledge. However, these existing approaches do not explicitly address the issue of missing measurements, either by excluding patients with any missing features [21], imputing missing values with median or recommended default [23], or no mention at all [22]. Furthermore, these deep learning architectures do not inherently support longitudinal data and do not incorporate real-time risk updates, which are crucial in the analysis of clinical data in real-world settings.

In light of the need for a model that can dynamically update risk assessment while being robust to varied data quality, our recent work investigated the performance of deep learning architecture known as gated recurrent unit with decay (GRU-D) proposed by Che et al [24] in risk assessment using real-world electronic health records (EHR). This architecture has demonstrated effectiveness in handling risk assessment tasks using longitudinal real-world data [25,26]. Building upon this, with CKD as a use case, we proposed a novel approach called GRU-D-Weibull [27], which leverages GRU-D to estimate the parameters of the Weibull distribution. The Weibull distribution is frequently employed for modeling event failure probabilities due to its versatility in representing various hazard function shapes with only two parameters. This characteristic makes it well-suited for diverse survival situations involving both early and late failures, such as in the case of CKD. With limited clinical features, GRU-D-Weibull exhibited state-of-the-art performance in CKD endpoint risk prediction, and addresses the challenges of longitudinal input, automated handling of missing data, and provides individualized survival distributions at any desired time point.

The primary aim of this study is to demonstrate the clinical utility of the GRU-D-Weibull architecture as a semi-parametric model for predicting composite endpoints in patients diagnosed with CKD stage 4 (CKD4). The study focuses on addressing the challenges associated with real-world data, including missingness, irregularity, and asynchronicity, which are commonly encountered in clinical practice. By incorporating comprehensive patient information, such as ICD/CPT codes and medication data, and a loss function that directly accounts for censored patients, our research showcases the enhanced prediction performance of the GRU-D-Weibull architecture. To evaluate the performance of our model, we compared it with GRU (without automated missing parameterization), Nnet-survival (a semi-temporal dense feed forward network), random survival forest (RSF), and traditional regression-based accelerated failure time (AFT) models with or without gradient boosting. Our results showed that GRU-D-Weibull constrained predicted survival probability at time of event within a remarkably smaller and more fixed range than the competing models, and achieved reasonably accurate point estimates over a prolonged follow-up period. The clinical utility of our approach was demonstrated by generating various endpoint estimates for real-time monitoring, and a simple post-training recalibration to align predicted probabilities with observed values. The findings underscore the potential of the GRU-D-Weibull architecture as a promising candidate for real-time individualized CKD4 endpoint risk management tools.

# Methods
## GRU-D-Weibull model architecture

The basic architecture of the GRU-D model has been systematically described by [24], here we only recapitulate the equations for handling missing values.

$$\hat{x}_t^d = m_t^d x_t^d + (1 - m_t^d)(\gamma_{x_t}^d x_{t'}^d + (1 - \gamma_{x_t}^d)\tilde{x}^d)$$

where $m_t^d$ is the missing value indicator for feature $d$ at timestep $t$. $m_t^d$ takes value 1 when $x_t^d$ is observed, or 0 otherwise, in which case the function resorts to weighted sum of the last observed value $x_{t'}^d$ and empirical mean $\tilde{x}^d$ calculated from the training data for the $d$th feature. Furthermore, the weighting factor $\gamma_{x_t}^d$ is determined by

$$\gamma_t = exp\{-max(0, W_\gamma \delta_t + b_\gamma)\}$$

where $W_\gamma$ is a trainable weights matrix and $\delta_t$ is the time interval from the last observation to the current timestep. When $\delta_t$ is large (i.e. the last observation is far away from current timestep), $\gamma_t$ is small, results in smaller weights on the last observed value $x_{t'}^d$, and higher weights on the empirical mean $\tilde{x}^d$ (i.e. decay to mean).

The GRU-D-Weibull architecture is shown in supplementary Fig 3. Specifically, the model is trained to minimize a composite loss function composed of a) negative log likelihood of Weibull's PDF (*neglog*) plus mean square log error of the difference between predicted median survival time (PMST) and observed remaining time to event (*MSLE*) (for uncensored patients), and b) cumulative probability of event from censoring to infinity (for censored patients)

$$loss_{total} = \sum_{i=1}^{N} \left( w_i \sum_{t=1}^{T_i} ((1 - C_i)(-1 * log(f(\tau_{it}, \lambda_{it}, \kappa_{it})) + MSLE(\tau_{it}, Median(\hat{\tau}_{it}))) + C_i(-1 * log(F(+\infty) - F(c_{it})))) \right)$$

(1)

equation (1) is composed of the following parts (subscripts $i$ and $t$ not displayed for clarity), where for uncensored patients

$$f(\tau; \lambda; k) = h(\tau)S(\tau) = \frac{k}{\lambda}\left(\frac{\tau}{\lambda}\right)^{k-1} e^{-(\tau/\lambda)^k} \quad (2)$$

is the probability density function of a 2-parameter Weibull distribution which models the $h(\tau)$ probability of event at time $\tau$. $\tau$ is the remaining time to event. Note $\tau$ decreases over time. is the unitarized hazard (i.e. probability of developing event in a unit of time) at time $\tau$. $S(\tau)$ is the probability of surviving up to time $\tau$.

$$Median(\hat{\tau}) = \lambda(-1 * log(0.5))^{(1/\kappa)} \quad (3)$$

is the PMST (i.e. probability of survival = 0.5).
For censored patients
$$F(c) = 1 - e^{-(c/\lambda)^\kappa}$$
$$F(+\infty) - F(c) = 1 - F(c) = e^{-(c/\lambda)^\kappa}$$
is the cumulative probability of events from the time of censoring $c$ to $\overline{y}$ (y upper bound, set to $+\infty$). $N$ is the number of patients in a training batch, $C_i$ is the censoring status (censored=1, uncensored=0) for patient $i$. $T_i$ is the timesteps before event or censoring for patient $i$ (i.e. no contribution to the model if the patient developed an endpoint or is censored). $w_i$ is individualized weights.

$$w_i = \begin{cases} 1, & \text{if not censored.} \\ t_c/5, & \text{otherwise, time of censoring divide by 5(yrs).} \end{cases}$$

---

**Algorithm 1:** Timestep sensitive composite loss based training of GRU-D-Weibull

---

Input (for patient $i$): $X_i = [x_i, m_i, d_i]$
 $x_i, m_i, d_i$ respectively are $f$ by $T_i$ tensors represent $f$ features with $T_i$ timesteps
Intermediate output: $\{[\kappa_{i0}, \lambda_{i0}]..[\kappa_{iT_i}, \lambda_{iT_i}]\}$
Final output: Average loss for a training batch
Initialize batch loss $Loss_{batch}$ to an empty array
**for** each patient $i$ in a training batch **do**
 Prepare weight for patient $i$ as $w_i = 1$ if not censored, or
  $w_i = Time\_of\_censoring/5(yrs)$
 **for** $t = \{1..T_i\}$ **do**
  $\kappa_{it}, \lambda_{it}$ = GRU-D( $x_{it}, m_{it}, d_{it}$ )
   Prepare $\tau_{it}$ as the remaining time to event and $c_{it}$ as the remaining time to censoring
   $Loss_{it}$ = WEIGHTED-COMPOSITE-LOSS($w_i, \kappa_{it}, \lambda_{it}, \tau_{it}$ )(equation 1)
   Append( $Loss_{batch}, WeightedLoss_{it}$)
 **end for**
**end for**
Calculate average batch loss $AvgLoss_{batch} = Average(Loss_{batch})$
Calculate model delta weight change BACKWARD( $AvgLoss_{batch}$)
Use gradient descent to update model parameters $W = W + \eta W$

---

The implementation of the GRU-D-Weibull model was performed on a unix system with NVIDIA Tesla V100 GPU. We reimplemented the original GRU-D code (in Keras format) associated with the original publication [24] to pytorch format (version 1.9.0) with python (version 3.8.10). The example code is available at github (https://github.com/xy-ruan/GRU-D-Weibull).

**Hyperparameters and training schemes**
The hyperparameters and training schemes were determined manually by varying combinations of neuron size (20-100), learning rate (0.001-0.1), batch size (30-500), drop out ratio (0-0.5), optimizer (SGD, RMSprop, ADAM), clipnorm (1,3), clipvalue (1,3) and early stopping threshold that optimize the performance of GRU-D-Weibull on 5000 samples (1879 held-out not touched). Finally, we found that a combination of 40 neurons, learning rate 0.001, 50 epochs, batch size 500, max 4% validation/training loss difference, and ADAM optimizer with amsgrad has overall better performance in terms of training speed and stability (e.g. resistance to gradient explosion).

**Competing models**
GRU-lvcf-Weibull analysis was implemented with pytorch (1.9.0) built-in GRU module. We performed unlimited timestep last value carry forward (LVCF) to fill in the missing measurement (mean and standard deviation calculated before LVCF for numeric variables) and trained with the same training scheme and loss function as GRU-D-Weibull. Specifically, the number of hidden neurons was increased to 80 to achieve a comparable number of trainable parameters with GRU-D-Weibull.

Nnet-survival was implemented with keras (2.4.0) with loss function proposed by [23], and assessed at discrete integer year time points ranging from -3 to 4 years, independently. The input value underwent the same LVCF processing as GRU-lvcf-Weibull. The model architecture consisted of a four-layer feed-forward network with an input layer, a hidden layer comprising 20 neurons, a dropout layer with 20% dropout ratio, and an output layer that predicted the conditional probability of survival for given time intervals. Notably, the input features were transformed into semi-temporal format by concatenating historical measurements. Various combinations of traceback period (including atemporal version that has no traceback) and sampling interval were explored, and the best performing configuration, featuring a maximum of 40 timesteps (~3 years) traceback and 4 timesteps (~120 days) sampling interval, was selected for further analysis.

For the accelerated failure time (AFT) model, XGBoost based AFT (XGB(AFT)) model, and random survival forest (RSF) model, a maximum 1-year LVCF approach was utilized for all input features. Subsequently, features with missing data exceeding 99.5% were excluded from further analysis. The remaining features underwent imputation using RSF adaptive tree imputation [28] with three iterations. Each of the three models were assessed at discrete integer year time points from -3 to 4 years, independently. Prior to fitting the AFT and XGB(AFT) models, binary features (such as CCS code, CPT code, administered and ordered medications) underwent collinearity elimination through backward selection using the R rms package. Only those features that demonstrated a significant reduction in the sum of square residuals were retained for further consideration. RSF analysis was conducted with R package randomForestSRC (version 3.1.1) with 100 trees and a minimum of 10 terminal nodes. The predicted survival probability for 1-5 years prediction horizons were retrieved by looking up the closest time grid (tolerance of +/-50 days). For non-parametric models (RST, Nnet-survival), the PMST was obtained through looking up the time grid where predicted survival probability was closest to 0.5 (tolerance of +/-0.1). For semi-parametric or parametric models (GRU-lvcf-Weibull, AFT, XGB(AFT)), the PMST was obtained through $\kappa$ and $\lambda$ according to equation 3.

**Dataset**
The aforementioned models were applied to CKD4 patients to predict the risk and time to composite endpoint defined as CKD stage 5, end stage renal disease (ESRD) or death [29], whichever occurs first. While the term "survival" usually means "time to death", throughout the paper we use "survival" interchangeably with "time to composite endpoint" for simplicity. The dataset includes a total of 9,479 patients with CKD4 diagnosis (ICD9 code 585.4 or ICD10 code N18.4) between 2005 and end of 2017 at Mayo Clinic Rochester, and was approved by the Mayo Clinic institutional review board (IRB number: 15-000105). All patients included in this study had consent for their data to be used for research. We excluded 1,862 patients with any kidney transplant record, or with CKD5, ESRD diagnosis outside of Rochester (to minimize the impact of lost follow-up or very sparse lab measurement of migrating patients). Afterwards, 738 patients with no lab record were removed. After exclusion, 6,879 patients were included in the final analysis (Fig 1), with baseline population characteristics shown in Table 1.

| Table 1 Baseline characteristic of study population | |
|---|---:|
| **Number of patients** | 6879 |
| **Age at CKD4 diagnosis** | 75(65,83) |
| **Gender** | |
| Male | 3743(54%) |
| Female | 3136(46%) |
| **Race** | |
| Caucasian | 6231(91%) |
| Non-caucasian | 648(9%) |
| **Follow-up time(years)** | 2.2(0.5,4.8) |
| **Censoring rate** | 3370(49%) |
| **Events type *** | |
| Death | 2663(76%) |
| CKD5 | 509(15%) |
| ESRD | 510(15%) |
| Data shown as n(%) or median(IQR) | |
| * 169 patients have CKD5 and ESRD on the same day. 2 patients have ESRD and death on the same day. 2 patients have CKD5 and die on the same day. | |

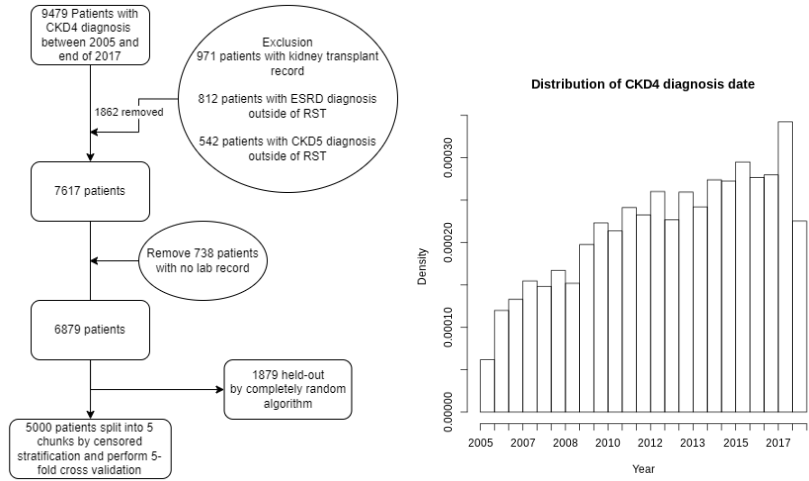

**Fig 1**  Sample preprocessing workflow and distribution of CKD4 diagnosis date

**5-fold cross validation and held-out split**

From the 6,879 qualified patients, 1,879 were selected through a completely random mechanism as a held-out dataset. For the remaining 5,000 patients, censored stratification strategy (put approximately the same number of uncensored observations in each fold but not pay any attention to event time) was used to split the patients into 5 chunks (n=1,000 per chunk). Notably, once the sample indices for the held-out and 5-fold datasets are determined, these indices are used consistently across all competing models for fair comparison. The detailed training, validation, testing, and held-out application strategies are shown in Supplementary Figure 4.

**Input features**

A total of 536 features including 532 dynamic (i.e. changes over time) and 4 static features were used in current analysis (Table 2). eGFR was estimated through the CKD-EPI equation [30]. For the GRU models, a non-uniformly distributed sampling scheme (every 15 days within half year around CKD4 diagnosis, and otherwise every 30 days) was adopted for dynamic features, based on the distribution of time intervals between measurements. This generated a total of 110 timesteps for each patient from 3 years before and up to 5 years after CKD4 diagnosis. Static features were replicated through timesteps. Continuous features like eGFR, SBP, BMI were z-score transformed based on mean and standard deviation (SD) of training dataset. Categorical variables like comorbidity and gender were binary encoded. For comorbidity and procedure codes we assume a 100-day effect window once a comorbidity is identified, and for medications a 30-day effect window once a medication record is identified, after which the effect is automatically handled by the missing value decay mechanism.

| Table 2 Features overview and preprocessing methods | | |
|---|---|---|
| | Preprocessing | Feature list |
| **Dynamic features** | | |
| Lab (numeric) | z-score | Estimated Glomerular Filtration Rate (eGFR), Blood Albumin, Phosphorus, Calcium, urine Albumin-to-Creatinine ratio (uACR), Bicarbonate, Creatinine |
| Vital (numeric) | z-score | Systolic Blood Pressure (SBP), Diastolic Blood Pressure (DBP) |
| Comorbidity (binary) | binary | 265 CCS code |
| Proceduce (binary) | binary | 244 CPT code |
| Medication (administered/Ordered treated separately) | binary | Angiotensin-Converting Enzyme Inhibitor (ACEI)/Angiotensin Receptor Blockers (ARBS), Benzodiazepine, Beta Blockers/Related, Calcium Channel Blockers, Combination Oral Pills, Insulin, Diabetes Oral Medications, Diuretics, Erythropoiesis-stimulating agents (ESA), Fibrates, Iodinated Contrast Media, Nonsteroidal anti-inflammatory drugs (NSAIDS), Statins |
| Others | see detail | Age [divide by 100], Body Mass Index (BMI) [z-score] |
| **Static features** | | |
| Demographics | binary | Gender, Race, Smoking, Alcohol |

### Diagnosis codes

ICD-9 diagnosis codes were cast into 265 categories of CCS (clinical classifications software) codes [31]. ICD-10 diagnosis codes were mapped to ICD-9 code through general equivalence mapping provided by Center for Medicare & Medicaid Services. CCS is a tool for clustering patient ICD-9 diagnoses and procedures into a manageable number of clinically meaningful categories for easy presentation and statistical analysis. The CCS diagnosis codes were then one hot encoded for the machine learning models.

### Healthcare Common Procedure Coding System (HCPCS)

HCPCS level I (Commonly known as Current Procedural Terminology (CPT) codes), HCPCS level II code, and ICD-9 procedure classification were cast into 244 CCS services and procedures code. ICD-10 procedure codes were first mapped to ICD-9 version through general equivalence mapping provided by Center for Medicare & Medicaid Services, and then cast into CCS services and procedures codes, which were then one hot encoded for the machine learning models.

### Medication

Drugs and corresponding categories were provided by nephrologists. A total of 197 drugs belonging to 12 categories are shown in Supplementary Table 1. The usage(or not) of drugs were binary coded for each of the 12 major categories. Specifically, routes relevant to ophthalmic or topical use were excluded due to anticipated very low systemic dosage. Combination drugs from different categories were independently identified and recorded for each category. For each drug, the length of use was retained and replicated throughout followup time. Drugs that were administered (i.e. ones given during hospitalization or at clinic) and ordered (i.e. for taking away) were coded separately as independent input features.

**L1-loss**
We calculated L1-loss for uncensored patients at each follow-up time point independently. L1-loss is defined as the average absolute value of the difference between PMST and observed time to event.

$$L1 = \frac{1}{|V_u|} \sum_{i \in |V_u|} |Median(\hat{\tau}_i) - \tau_i|$$

where $V_u$ is the subset of uncensored patients that have not reached the endpoint yet at corresponding time point.

**Calibration**
Calibration was measured by splitting ascendingly ordered predicted survival probability (i.e. $S(\tau) = exp[-(\tau/\lambda)^k]$) into 10 chunks, and for each chunk comparing the average predicted versus observed survival probability.

**Results**
**C-index throughout follow-up time**
Among the competing models, namely GRU-D-Weibull, AFT, XGB(AFT), and RST, a notable enhancement in the C-index was observed after the CKD4 index date compared to the pre-index period (Fig 1). GRU-D-Weibull exhibited the optimal C-index for prediction horizons ranging from 3 to 5 years, with an improvement to ~0.7 at the index date, and ~0.74 in the 5-fold cross-validation as well as ~0.77 (+/-0.02 95% CI) in the held-out after a follow-up period of approximately 4 years. Note that GRU-D-Weibull has a relatively diversified C-index across prediction horizons due to individualized shape parameter $\kappa$ for each patient at each time point of follow-up. As a comparison, competing models including AFT, XGB(AFT), and RSF also have C-index around 0.7 at index date. RSF achieved a maximum of 0.75 (+/-0.014 95% CI) at 3 yrs of follow-up. Notably, AFT and XGB(AFT) have the same C-index on 1-5 year prediction horizons due to the assumption of fixed shape parameter $\kappa$ for all patients at a given time point. Although GRU-lvcf-Weibull was trained with comparable configurations (including optimizer, cross-validation/held-out split, overfit early stopping threshold, loss function, etc.) as GRU-D-Weibull, it exhibited a failure to effectively capture the relationship between input features and the outcome. On the other hand, Nnet-survival with semi-temporal input features demonstrated a similar C-index to GRU-D-Weibull at the index date; however, its performance notably declined during the follow-up period after the index date.

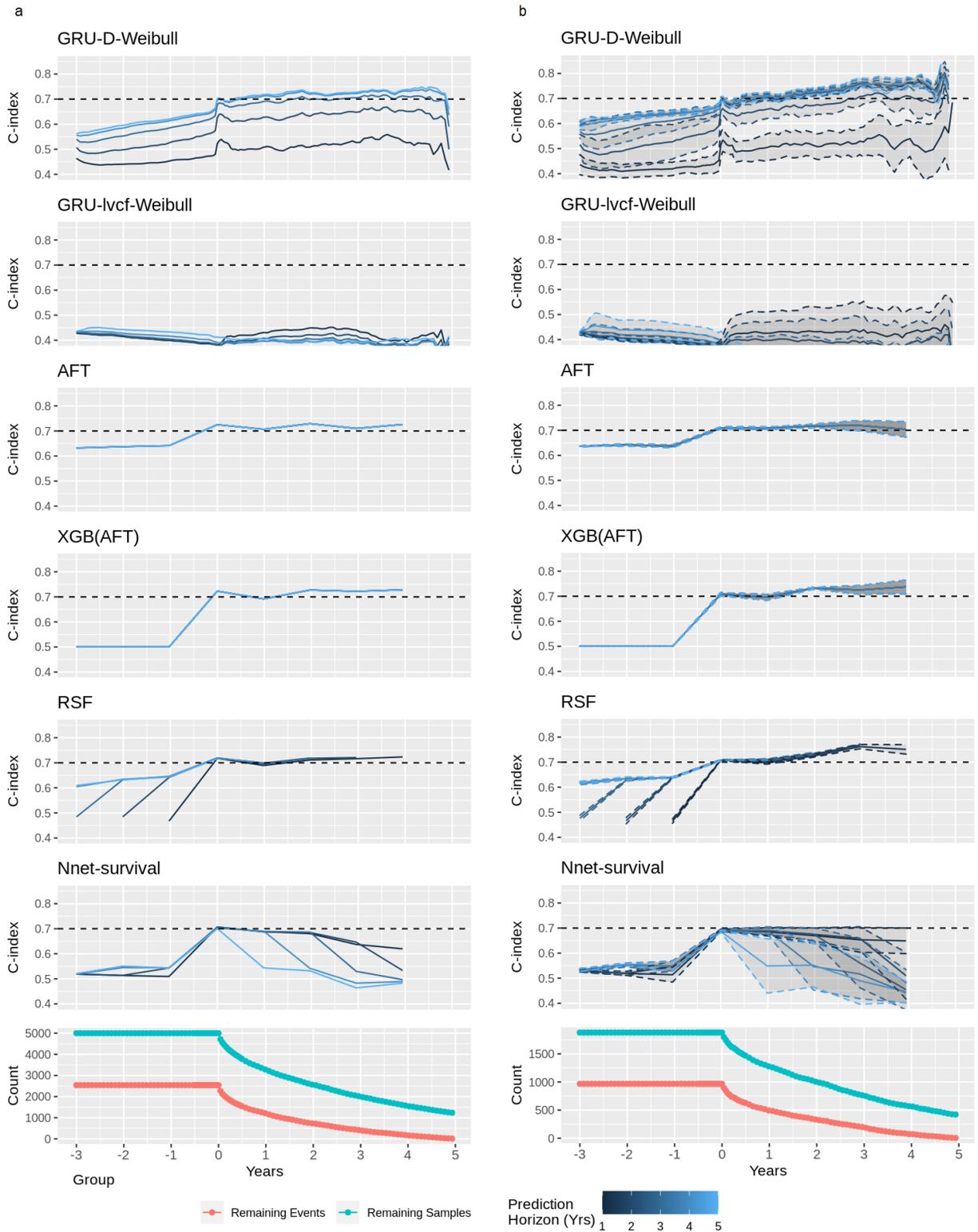

**Fig 1** C-index of competing models with predicted survival probability at 1 to 5 years prediction horizon in a) 5-fold cross validation and b) held-out dataset. Dashed black

line marks C-index equals to 0.7 for easy comparison. Ribbons are 95%CI obtained by applying each of the 5-fold models to the held-out.

**L1-loss throughout follow-up time**
We compared L1-loss of competing models to gain an intuitive understanding of how point estimates of survival time deviate from observed time to event at each time point of follow-up. For AFT and XGB(AFT) models, the L1-loss for uncensored patients at index date are generally similar to previously reported [27] (about 3 years for AFT and 1.5 yrs for XGB(AFT)). Both models have L1-loss and standard deviation (SD) deteriorated remarkably after the index date. GRU-lvcf-Weibull and RSF have lower (better) L1-loss than GRU-D-Weibull from 3 to 2 year(s) before index date (Table 2). GRU-lvcf-Weibull quickly deteriorated after the index date. As a comparison, GRU-D-Weibull, RSF, and Nnet-survival have continuously improved L1-loss and SD throughout the follow-up period (Fig 2). Among all competing models, GRU-D-Weibull has the lowest L1-loss and SD from index date to the end of 5-year follow-up, and reached ~0.5 year absolute L1-loss at 3 to 4 years after index date (Table 2).

a 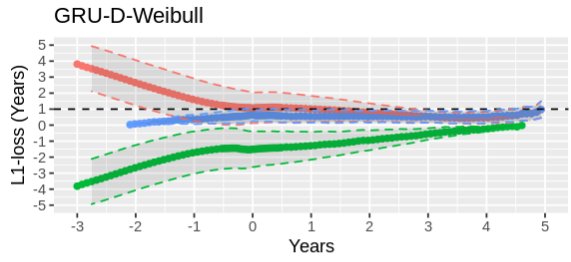 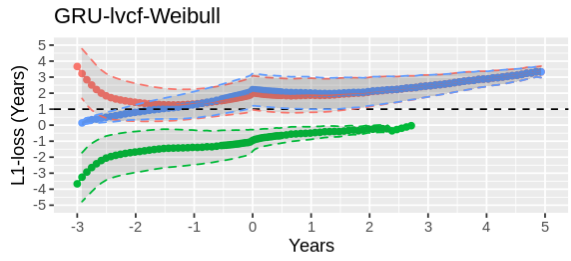 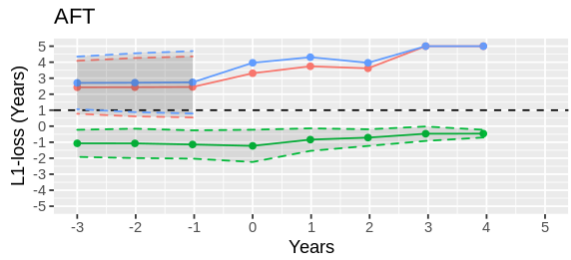 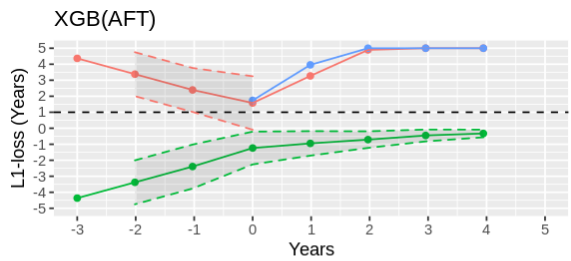 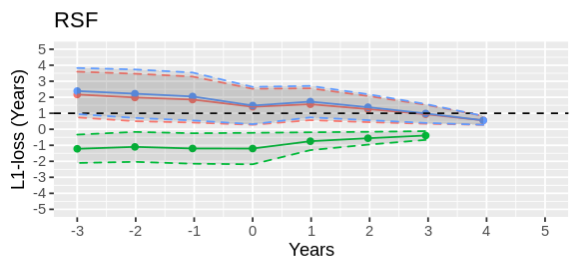 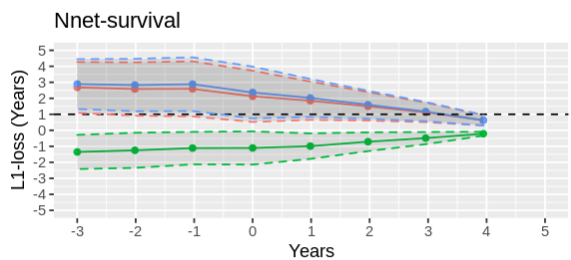

b 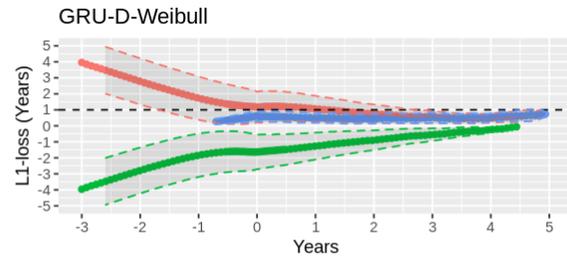 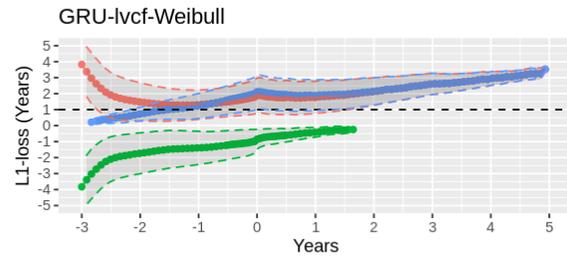 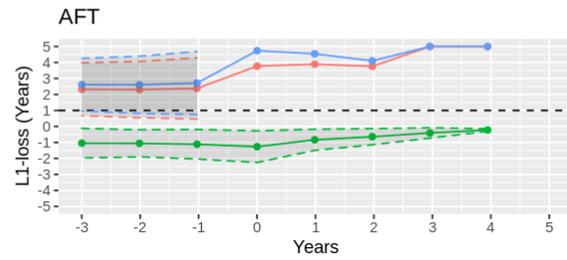 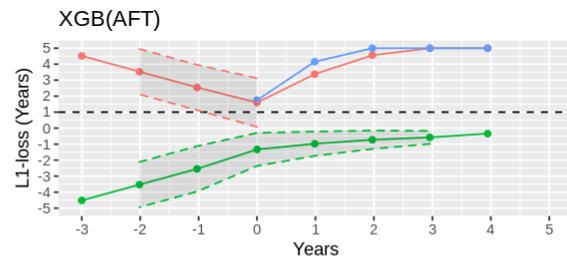 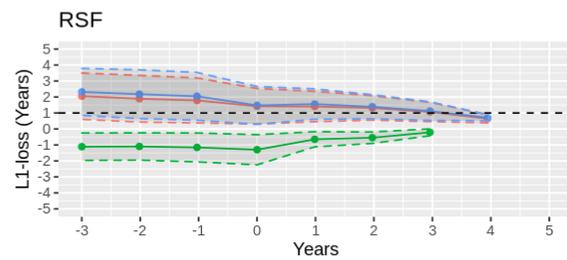 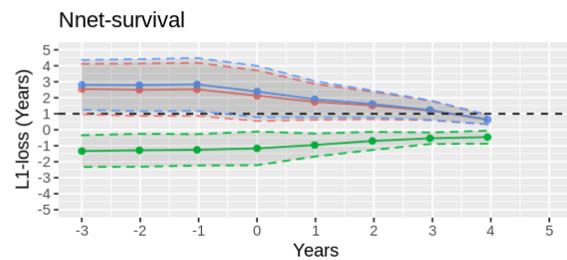

**Fig 2** L1-loss of uncensored patients based on differences between PMST and observed survival time for a) 5-fold cross validation and b) held-out dataset. The dashed black line marks 1 year prediction error for easy comparison. The red, blue, and green curves and ribbons are L1-loss based on absolute value, only positive values (overestimation) and only negative values (underestimation), respectively. For the held-out dataset, data shown are average L1-loss and average SD obtained through 5 cross-validation models. For AFT and XGB(AFT), SD values greater than 5 years were not displayed for clarity.

Analysis of the proportion of serious error by Parkes' definition (i.e Target > 2*Prediction or Target < 0.5*Prediction) indicated that GRU-lvcf-Weibull, AFT, RSF, and Nnet-survival have generally lower error proportion before index date. The six competing models performed similarly at index date, after 1-year of follow-up GRU-D-Weibull has predominantly lower error proportion and less deviation from tolerable prediction window (i.e. > 0.5*Target and <2*Target) (Sup Fig 1). The lowest error proportion was recorded by GRU-D-Weibull with 41% and 42.6% (+/-6% 95%CI) at 3.5 years for 5-fold cross validation and held-out, respectively.

Table 2 Average absolute L1-loss throughout follow-up time for uncensored patients

| L1-loss (Mean, Median, SD) | -3 yrs | -2 yrs | -1 yr | Index date | 1 yr | 2 yrs | 3 yrs | 4 yrs |
|---|---|---|---|---|---|---|---|---|
| **5-fold cross validation** | | | | | | | | |
| GRU-D-Weibull | 3.81, 3.33, 1.38 | 2.66, 2.27, 1.42 | 1.61, 1.15, 1.33 | **1.06, 0.75, 0.93** | **1.01, 0.79, 0.82** | **0.79, 0.68, 0.59** | **0.56, 0.49, 0.38** | **0.45, 0.35, 0.36** |
| GRU-lvcf-Weibull | 3.67, 3.2, 1.44 | **1.43, 1.07, 1.18** | **1.32, 1.2, 0.93** | 2.03, 2.09, 1.07 | 1.87, 1.93, 1.07 | 2.06, 2.14, 0.92 | 2.44, 2.44, 0.7 | 2.88, 2.94, 0.49 |
| AFT | 2.43, 2.14, 1.66 | 2.44, 2.11, 1.82 | 2.45, 2.0, 1.91 | 3.31, 1.39, 6.2 | 3.74, 2.15, 4.74 | 3.62, 2.48, 4.0 | 5.09, 3.18, 7.34 | 6.17, 3.74, 7.83 |
| XGB(AFT) | 4.36, 3.88, 1.37 | 3.38, 2.9, 1.37 | 2.39, 1.91, 1.37 | 1.58, 1.05, 1.67 | 3.27, 1.71, 4.58 | 4.9, 1.6, 17.35 | 53.56, 2.04, 495.27 | 7.12, 3.41, 9.98 |
| RSF | **2.17, 1.92, 1.43** | 1.99, 1.58, 1.48 | 1.86, 1.5, 1.44 | 1.41, 1.1, 1.12 | 1.56, 1.46, 1.0 | 1.26, 1.18, 0.82 | 0.94, 0.91, 0.59 | 0.56, 0.57, 0.28 |
| Nnet-survival | 2.68, 2.68, 1.59 | 2.58, 2.44, 1.67 | 2.59, 2.5, 1.72 | 2.12, 1.85, 1.6 | 1.85, 1.76, 1.2 | 1.5, 1.43, 0.89 | 1.12, 1.15, 0.61 | 0.63, 0.65, 0.33 |
| **Held-out** | | | | | | | | |
| GRU-D-Weibull | 3.96, 3.47, 1.42 | 2.8, 2.36, 1.44 | 1.75, 1.27, 1.36 | **1.16, 0.82, 0.97** | **1.07, 0.87, 0.82** | **0.76, 0.61, 0.59** | **0.56, 0.5, 0.39** | **0.45, 0.44, 0.27** |
| GRU-lvcf-Weibull | 3.83, 3.33, 1.46 | **1.49, 1.09, 1.21** | **1.3, 1.2, 0.92** | 1.9, 1.98, 1.08 | 1.76, 1.73, 1.01 | 2.12, 2.23, 0.87 | 2.6, 2.67, 0.68 | 2.9, 2.95, 0.43 |
| AFT | 2.32, 1.99, 1.65 | 2.3, 1.92, 1.76 | 2.37, 1.95, 1.91 | 3.77, 1.51, 7.97 | 3.89, 1.93, 5.94 | 3.76, 2.83, 3.46 | 5.6, 3.38, 8.98 | 6.77, 4.73, 7.0 |
| XGB(AFT) | 4.52, 4.08, 1.42 | 3.53, 3.09, 1.42 | 2.54, 2.11, 1.42 | 1.6, 1.14, 1.52 | 3.38, 1.63, 6.81 | 4.56, 1.75, 10.4 | 55.9, 2.4, 257.06 | 53.55, 4.07, 403.09 |
| RSF | **2.05, 1.74, 1.45** | 1.89, 1.48, 1.46 | 1.78, 1.38, 1.41 | 1.42, 1.14, 1.11 | 1.4, 1.25, 0.95 | 1.31, 1.24, 0.77 | 1.06, 1.15, 0.61 | 0.64, 0.73, 0.26 |
| Nnet-survival | 2.54, 2.44, 1.58 | 2.5, 2.3, 1.64 | 2.52, 2.35, 1.66 | 2.13, 1.85, 1.59 | 1.74, 1.63, 1.13 | 1.51, 1.53, 0.86 | 1.2, 1.25, 0.62 | 0.63, 0.65, 0.29 |

**Predicted survival probability at time of event**

In the analysis of uncensored patients, we examined the distribution of predicted survival probabilities at the time of event, specifically focusing on integer years of follow-up and all time points subsequent to the index date. The distribution, as depicted in Fig 3, revealed that the majority of patients in the GRU-D-Weibull model exhibited estimated survival probabilities centered around 0.25 at the time of event, consistently observed throughout the entire follow-up period. In comparison, the competing models demonstrated estimated survival probabilities that were generally between 0.75 and 1.0 at the time of event, exhibiting a wide distribution ranging from 0 to 1. Notably, for all models, the predicted survival probability exhibited an increasing trend over the follow-up period, with Nnet-survival showing the most pronounced and extreme increases.

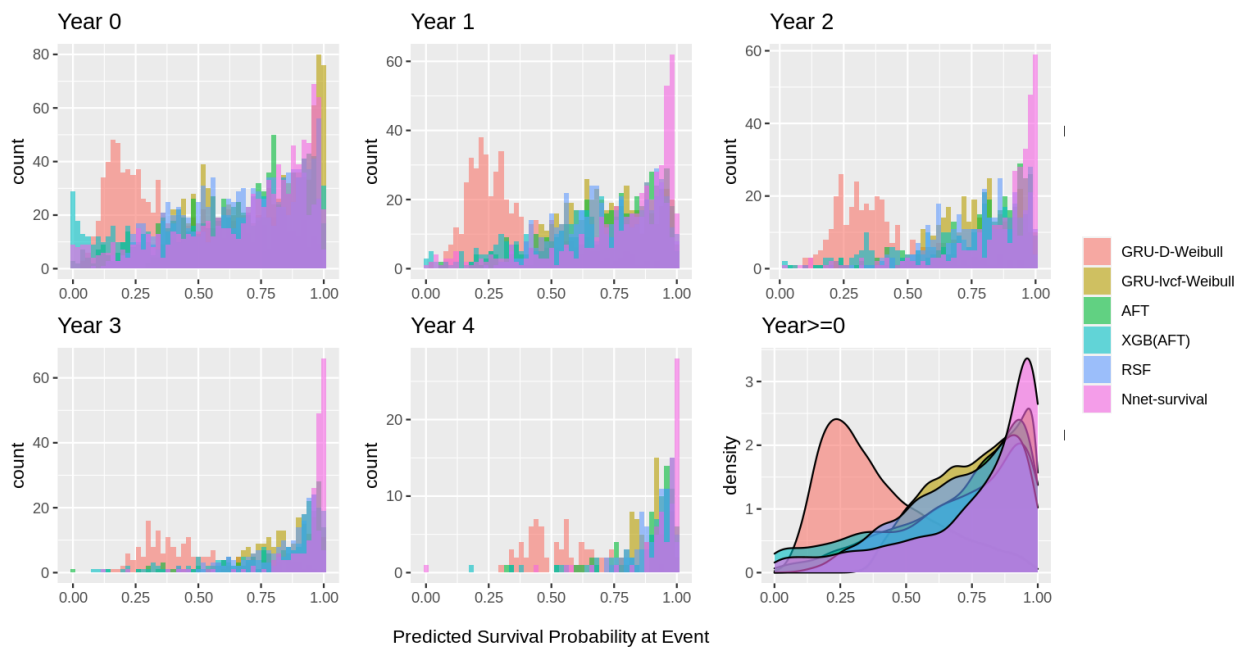

**Fig 3** Distribution of predicted survival probability at time of event for uncensored patients at 0 to 4 years of follow-up, and all time points after index date (Year >= 0). Data shown for the held-out dataset predicted with one of the 5-fold models. Other folds have similar results.

**Prediction error and missing proportion**

Using the held-out dataset, we conducted an analysis to assess the prediction errors of GRU-D-Weibull, measured by the absolute value of Log(Target/PMST), and their correlation with the proportion of missing data across five major categories of input features (Fig 4). We also performed a similar analysis using Abs(Log((Target + 0.1)/(Prediction + 0.1))) to mitigate the impact of large ratios resulting from small numerator/denominator values (figure not shown). The results revealed a significant positive correlation between prediction errors and the proportion of missing data, primarily observed for predictions made at the CKD4 index date. This correlation gradually diminished over the follow-up period as more data became available. Among the input features, the strongest correlation was observed with the absence of administered/ordered drug usage immediately before the index date (corr=0.25 (+/-0.06), p<0.001). Additionally, a notable correlation was observed for missing dynamic features approximately 1 year prior to the index date (corr=0.13 (+/-0.06), p<0.001). We also observed similar but weaker correlations for ICD and CPT codes.

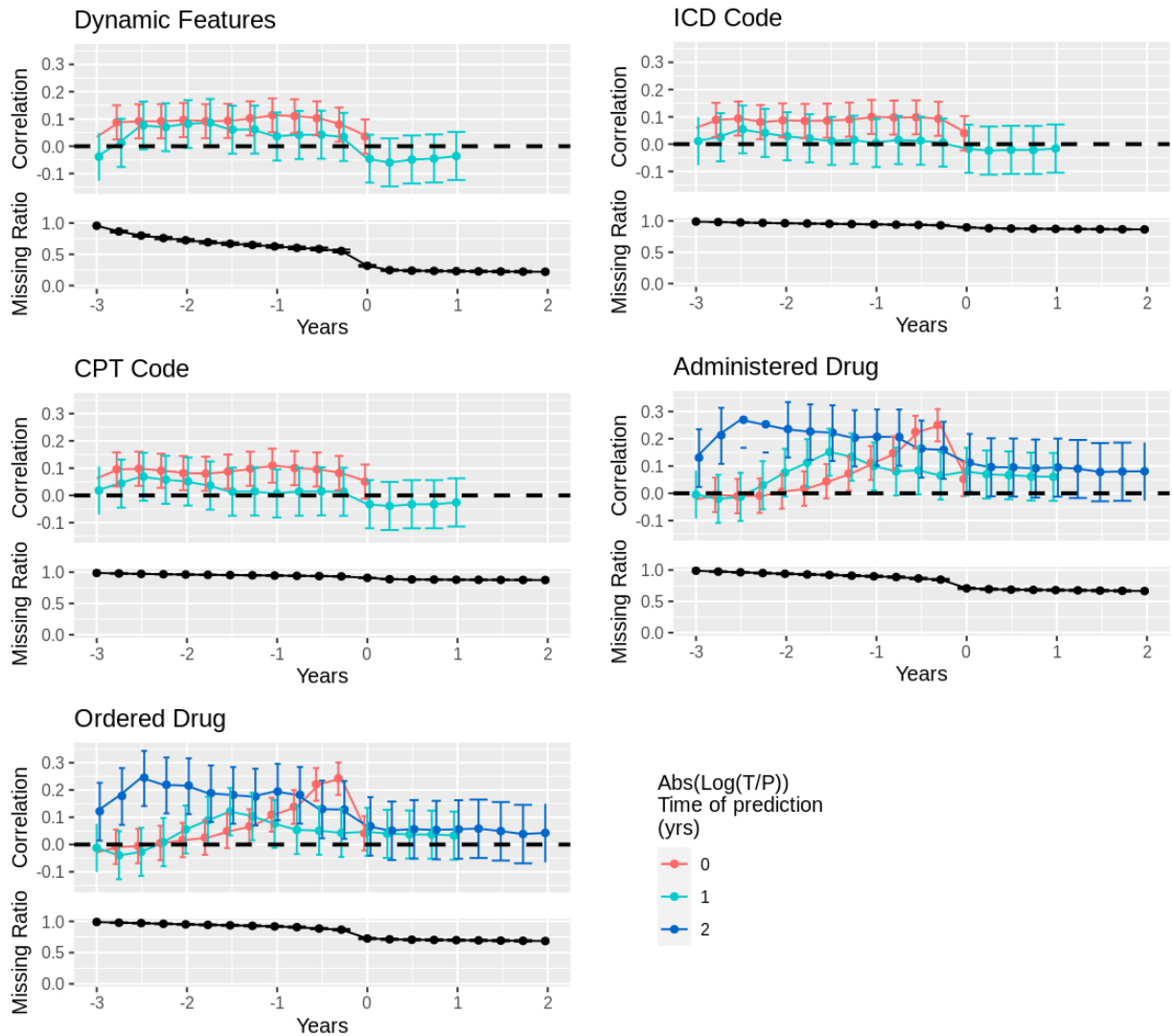

**Fig 4** Correlation of prediction error (measured by Abs(Log(Target/PMST))) at 0,1,2 years of follow-up with missing (Dynamic/ICD/CPT) or absence of drug usage proportion (Administered/Ordered Drug) at each time point before making prediction. All features processed with unlimited LVCF before calculating missing or drug usage proportion.

## Illustration of individualized endpoint prediction

In order to evaluate the individualized endpoint prediction performance of GRU-D-Weibull, we randomly selected 50 uncensored held-out patients and presented the results in Fig 5. This figure illustrates the Log(T/P) (where T represents the target and P represents the prediction) and baseline-adjusted 1-year cumulative hazard, with reference points set at -3 years and -1 year. The following observations were made: 1) Point estimates given by the PMST appear to be reasonably accurate within a time window of approximately 1.5 to 3 years prior to reaching the endpoint. 2) Utilizing the predicted survival time at 25% survival probability (as suggested by Fig 3) yields a notably longer period of accurate point estimates, albeit with some overestimation near the endpoint. 3) Overestimation of survival time, indicated by the red color in the Log(T/P) plot, predominantly occurs in proximity to the endpoint and exhibits a consistent pattern across patients with varying lengths of survival. 4) Approximately one third of the patients experience a sudden increase in the cumulative hazard within the year preceding the endpoint.

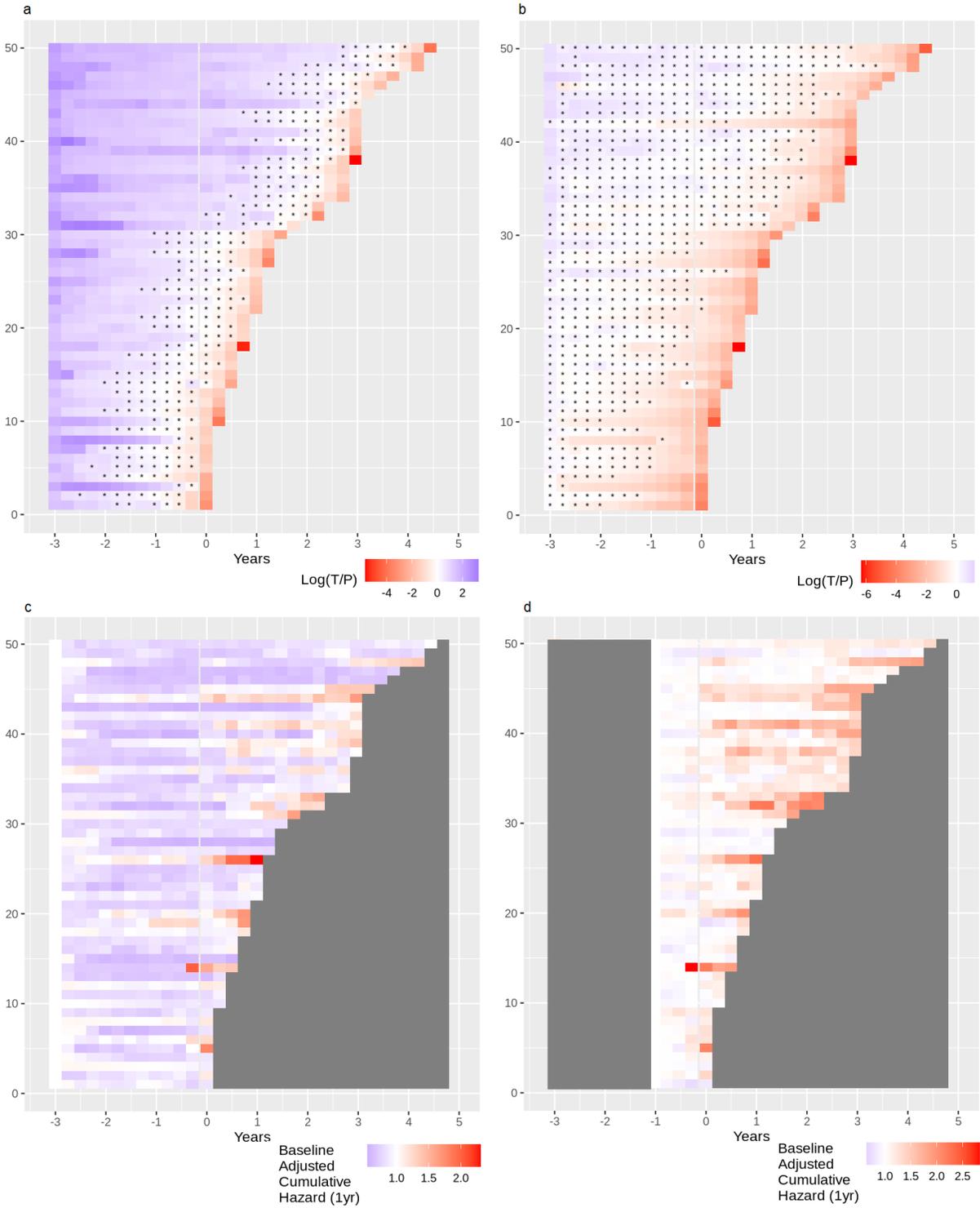

**Fig 5** Illustration of real-time endpoint prediction for 50 randomly selected uncensored held-out patients. Each tile spans ~90 days. a) prediction error measured by Log(Target/PMST), where red and purple colors represent overestimation and underestimation of true survival time, respectively. An asterisk within each tile indicates

the Target/Prediction falling within the Parkes' serious error boundary. b) displays the Log(Target/Predicted Survival time at 25% survival probability. Figure c) and d) showcase the baseline adjusted 1-year cumulative hazard, where the baseline is the 1-year cumulative hazard measured at c) -3 years and d) -1 year for corresponding patients, respectively.

In Fig 6, we presented two uncensored patients (randomly selected from the held-out dataset) with endpoints occurring at 687 days (patient 1) and 530 days (patient 2), respectively, highlighting various metrics derived from the GRU-D-Weibull architecture. These metrics closely resemble an individualized real-time endpoint risk management system applicable in clinical practice. Throughout the follow-up period, we observed the following trends: 1) a decreasing scale parameter, PMST, and survival probability at different prediction horizons. 2) an increasing instantaneous and cumulative hazard, with a notable spike in hazard near the endpoint. 3) patient 2, who had a shorter survival time compared to patient 1, exhibited generally lower lambda (scale parameter), PMST, survival probability, and higher hazard. These metrics provide valuable insights into the individualized risk profiles of patients and can inform clinical decision-making processes.

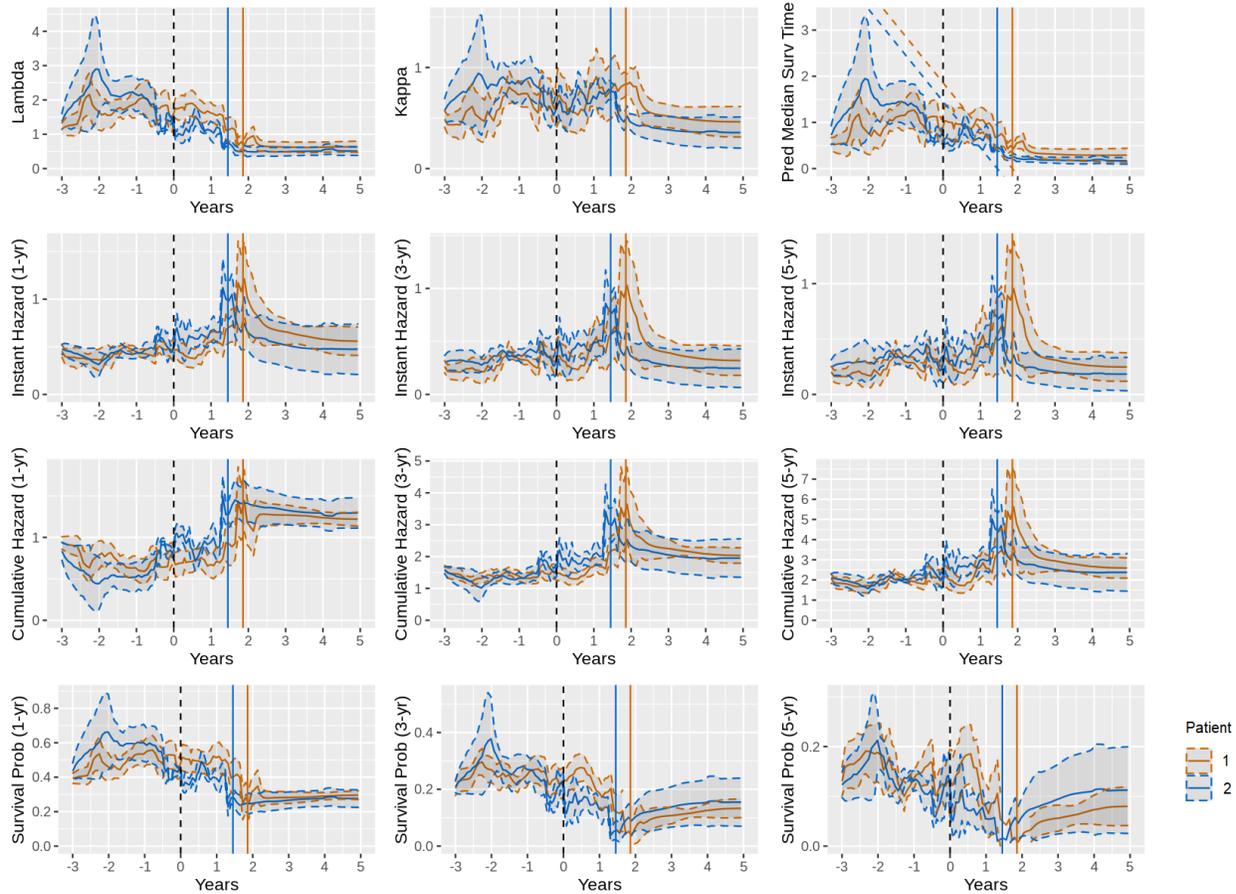

**Fig 6** Illustration of individual level endpoint prediction. Vertical brown and blue lines mark the observed endpoint of patient 1 (687 days) and 2 (530 days), respectively. The dashed backward slash lines in predicted median survival time plot mark the actual remaining time to endpoint. Ribbon areas represent 95% CI from 5 models trained with 5-fold cross validation.

**Calibration**

At the population level, the original predictions generated by the GRU-D-Weibull model tend to underestimate the true survival probability. For instance, when considering a 2-year prediction horizon, the group of patients with a predicted survival probability of 20% actually has an observed survival rate of 30% (Fi7 6a). It is worth noting that similar linear relationships with comparable slopes are observed across multiple follow-up time points within the same prediction horizon, albeit with slight variations (Fig 7a, b). To address this underestimation, we fitted a single linear regression model on the cross-validation dataset, obtaining the regression coefficient ($\beta \approx 2$). Subsequently, we demonstrated that by applying this recalibration technique to the held-out dataset, the predicted survival probability can be adjusted to align more closely with the observed probability (Fig 6c). In practice, separate regression lines can be fitted for different prediction horizons to achieve more accurate recalibration results.

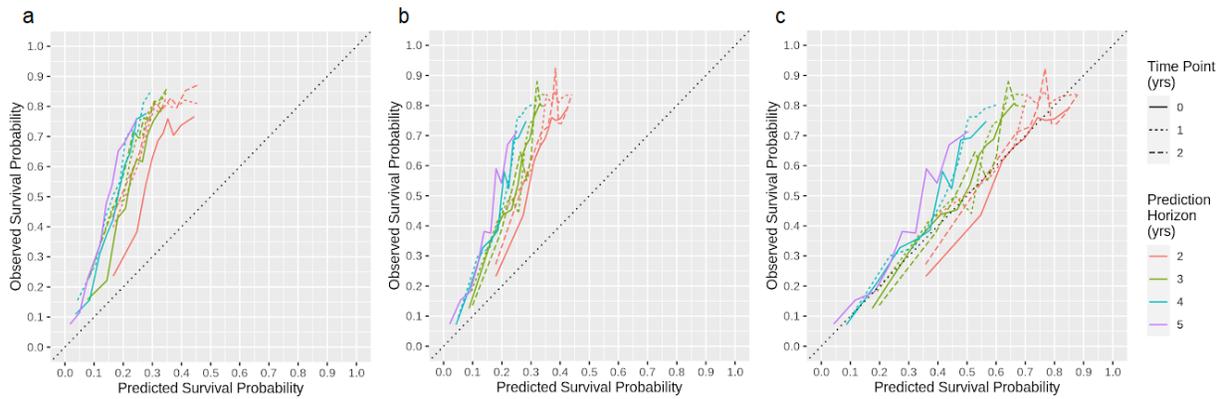

**Fig 7** Comparison of predicted versus observed survival probabilities at index date and 1, 2 years of follow-up across the 2 to 5 years prediction horizon. The comparisons are shown for a) cross-validation dataset, b) held-out dataset (illustrated with one of the five trained models), and c) held-out dataset recalibrated using a single linear regression model fitted on the cross-validation dataset

**Discussion**

We previously explored [27] various performance metrics of GRU-D-Weibull architecture in comparison with several atemporal models, and documented approximately one year prediction error after CKD4 index date. In this study, using the same study cohort and training/testing split but incorporating more comprehensive clinical features, we demonstrate that the point estimates based prediction error can be further reduced to a minimum of ~0.5 year within a 5-year follow-up period. In contrast, regression-based models such as AFT and XGB(AFT) exhibit significantly poorer accuracy in point estimates due to their reliance on population estimates, making them unsuitable for individualized predictions. While point estimates of survival time are useful for intuitive understanding and comparison across models/settings, they may be largely inaccurate for individual patients due to inherent uncertainty, as argued by previous studies [32] [33] [34]. Therefore, instead of reducing survival probability to a single point estimate, we emphasize the importance of constraining the predicted probability of survival at the time of the event within a relatively small and fixed range, which is a more practical and meaningful property of the GRU-D-Weibull architecture. The expression such as "the model exhibits a confidence level of *c*% in predicting the endpoint occurrence between *a* and *b* years" offers a more practical and applicable representation of predictions, while allowing for the exploration of hyperparameters (i.e., a, b, c) across diverse practical scenarios. To advance the practical implementation of the proposed model, future studies are needed by integrating it into clinical workflows, actively involving healthcare practitioners and assessing its impact on decision-making and patient outcomes. Specifically, there is a need to explore how the multiple outputs of the model, such as instant/cumulative hazard, survival probability, and point estimates of time to endpoint at various time points during follow-up, can be translated into actionable intervention strategies. Thorough validation studies are essential to refine the architecture, optimize its effectiveness in enhancing patient care, and offer valuable insights for its practical application in real-world clinical practice.

Among the models evaluated, only RSF exhibited competitive performance in terms of both C-index and L1-loss, particularly when dealing with highly sparse input features in the 3 to 1 year period before the index date. We note that the current RSF framework may benefit from a more intricate design, such as a dynamic RSF framework [35], to achieve similar utility as the GRU-D-Weibull architecture. One area for improvement is constraining the predicted survival probability at the time of the event to a narrower range, as the current RSF predictions tend to overestimate and exhibit wide spread between 0 and 1. Another limitation of RSF as a fully nonparametric model is that it only estimates survival probabilities at the specific time points where events or censoring occur in the training data. Consequently, point estimates can only be approximated based on the nearest available survival probability, provided it is close enough (within $\pm 0.1$ in current study), or otherwise unavailable.

The improvement in C-index throughout follow-up suggests that the GRU-D-Weibull model has the potential to provide enhanced risk assessment at the population level by incorporating newly available measurements. Interestingly, when comparing the standard GRU model (without a missing decay mechanism) to our proposed model, we found that the naive LVCF (GRU-lvcf-Weibull) was not a feasible option in the context of high missing data scenarios. Similarly, our investigation of the Nnet-survival model revealed that concatenating historical measurements as semi-temporal inputs and training with a dense feed-forward network did not yield satisfactory performance after the index date. Furthermore, we explored an approach involving limited time span (maximum of one year) LVCF followed by mean value carry forward, but this did not result in any improvement (results not presented). These findings highlight the importance of the delta time based missing decay mechanism in the GRU-D model for accurately capturing longitudinal measurements in EHR data with high levels of missingness. Future advancements could potentially be achieved by incorporating variable-sensitive GRU-D model [36], which considers feature-specific missing proportions.

Our findings revealed a notable correlation between the point estimates errors and the proportion of missing data, aligning with the intuitive understanding that a more comprehensive collection of information leads to improved prediction accuracy. Notably, the absence of drug usage information exhibited the most substantial impact, highlighting the significance of medical interventions on the survival trajectory of individuals with late-stage CKD [37]. As the correlation diminished one year after the index date, coinciding with the availability of additional data, we anticipate that enhanced data availability prior to the index date could further enhance prediction accuracy. These results underscore the importance of comprehensive data collection and suggest that efforts to augment data completeness could yield improved predictive performance in CKD patients.

For censored patients, we adopted a loss function maximizing the probability of surviving up to the time of censoring, as mentioned by [38], which makes the current training workflow more independent and integrated than the previous report [27]. However, it is mathematically provable this may cause shape parameter $\kappa$ to approach 0 (Sup Fig 2). This partly explains the lower C-index on 1-year prediction horizon, as there is more variation to the lower end of cumulative Weibull distribution when $\kappa$ is small. Setting $\bar{y}$ to a fixed value (e.g. 10 years) or a dynamic value (e.g. approaching 5 years during follow-up), or fix $\kappa$ to a constant (e.g. $\kappa = 1$) effectively make C-indices more consistent on multiple prediction horizons, however, at the cost of overall concordance and other metrics (data not shown). Despite this, under current training settings the predicted survival probability at 3-5 years prediction horizons provides a consistent and optimal estimate of C-index, and can be used with priority.

In our experiment of a loss function with an additional expected calibration error (ECE) [39] term, calibration can only be achieved at significant cost of C-index, L1-loss, and training time (data not shown). Our analysis indicated post-training recalibration as a much more viable option in terms of both computational efficiency and calibration performance for the current scenario. However, we do not rule out that other calibration [40] algorithms may provide better cost-performance value. A natively well-calibrated GRU-D-Weibull without significant hike in training time and comparable performance metrics could be an elegant target for future research.

**Conclusion**
We demonstrated the clinical utility of the GRU-D-Weibull deep learning architecture as a semi-parametric model for predicting composite endpoints in CKD4 patients. This architecture holds significant potential for diverse applications. The use of point estimates of time-to-event improves patient comprehension of risk, while healthcare practitioners can optimize risk management by comparing survival probabilities among patients. The recalibrated predicted survival probability facilitates accurate doctor-patient interactions. Importantly, the same architecture can be applied to similar scenarios, offering built-in parameterization of missingness and the ability to output both probabilities and point estimates at arbitrary time points and prediction horizons. With its capacity to handle sparsity, asynchronicity, individual uncertainty, and real-time evaluation demands in clinical practice, the GRU-D-Weibull model shows promise as a next-generation endpoint risk management tool.

**Limitations**
Our analysis demonstrates that the completeness of the longitudinal data impacts model performance. More investigation is needed to systematically assess the impact. We also did not systematically assess model fairness (e.g. comparable performance of C-index, L1-loss, and calibration in different gender, race, age groups, etc.) due to the lack of representativeness in the cohort. A detailed solution is out of the scope as the study here is to demonstrate the clinic utility of GRU-D-Weibull architecture. Even the proposed architecture factors in the explainability but more research is needed to understand questions like why a spike in instant hazard is observed for a patient at specific time of followup. Finally, future investigations should actively involve clinicians to gain a comprehensive understanding of the clinical utility of the GRU-D-Weibull model in individual patient care and validate its performance in diverse clinical scenarios. Particularly, it is important for future studies to include a thorough investigation on how continuous prediction updates can support clinical decision-making and financial preparedness at the individual patient level, as late-stage kidney disease often necessitates extensive medical interventions and ongoing treatments, which can have financial implications for patients and their families. For example, providing examples of how the ability to adapt predictions based on updated information becomes crucial in the management of chronic diseases like kidney disease, cardiovascular conditions, or cancer, where a patient's health status can change over time. Active involvement of clinicians will enable the collection of their insights and an assessment of the practical implications associated with implementing the GRU-D-Weibull architecture as an endpoint risk management tool in personalized clinical practice. By exploring the integration of our predictive model into clinical workflows and assessing its impact on decision-making and patient outcomes, valuable insights can be obtained to guide its practical implementation in real-world clinical practice.

## Acknowledgement

This work was supported by National Institute of Biomedical Imaging and Bioengineering grant R01 EB019403

## Supplementary Figures

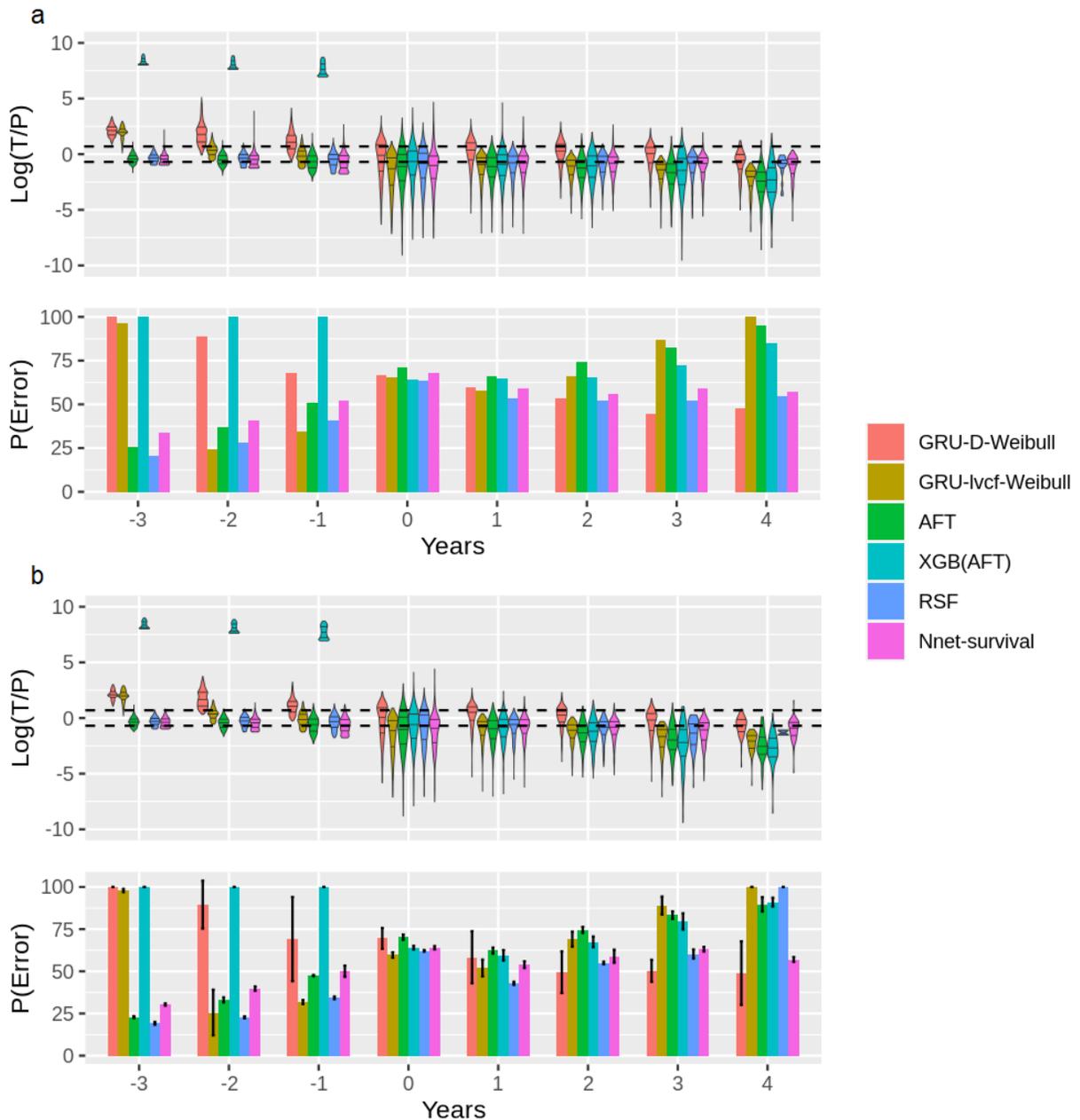

**Sup Fig 1** Distribution of Log(Target/Prediction) and proportion of serious error by Parkes' definition for a) 5-fold cross validation and b) held-out dataset. Dashed horizontal black lines mark the location of $\pm log(2)$. For the held-out dataset, average predictions from the 5 trained GRU-D-Weibull models are used for violin plot. The error bar represents the confidence interval of 5 trained GRU-D-Weibull models.

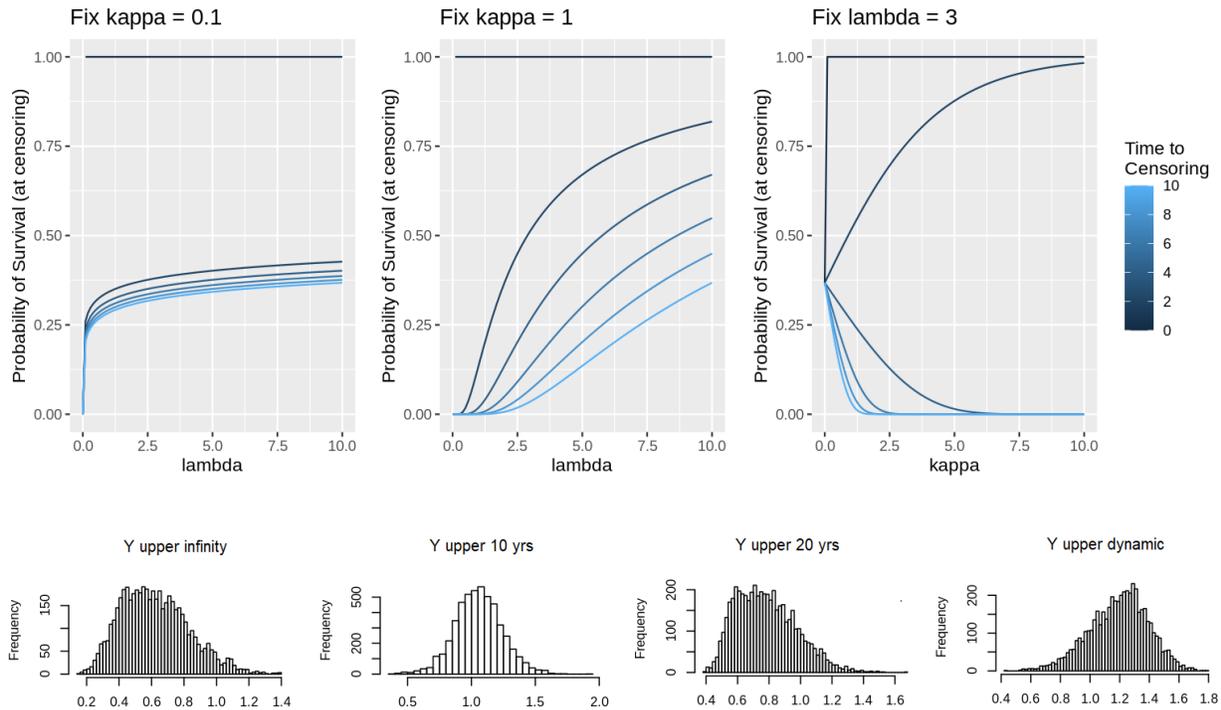

**Sup Fig 2**  a) The scale parameter $\lambda$ and shape parameter $\kappa$ that maximize probability function $F(+\infty) - F(c_{it}) = e^{-(c_{it}/\lambda)^\kappa}$ for censored patients, assuming $\bar{y}$ (Y upper bound) equal to positive infinity. Note $\kappa$ approaches either 0 or positive infinity depends on the ratio of $c_{it}/\lambda$. b) Distribution of $\kappa$ when $\bar{y}$ is positive infinity, fixed 10 years, fixed 20 years, and dynamically determined (assume all censored patients have an endpoint at 5 years).

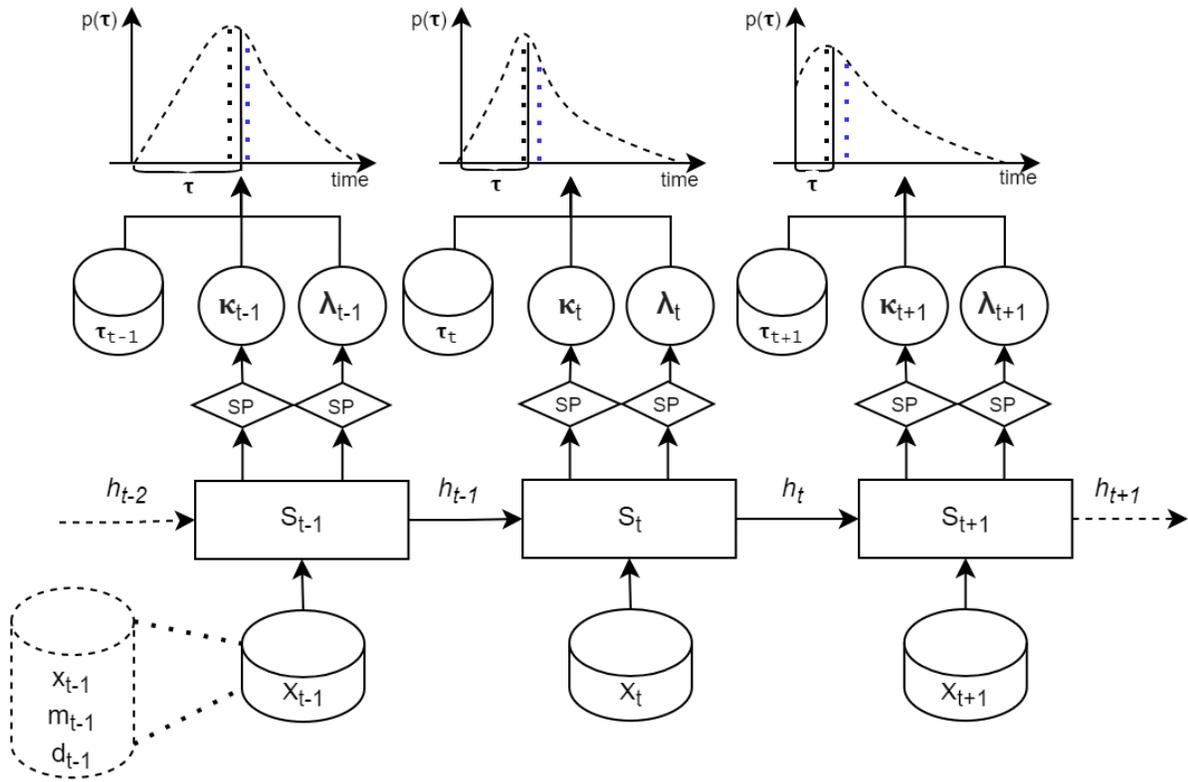

**Sup Fig 3**  GRU-D architecture that outputs 2-parameter Weibull's distribution. The shape parameter $\kappa$ and scale parameter $\lambda$ are updated at each timestep to maximize composite loss of negative log likelihood Weibull's PDF plus MSLE between predicted median survival time (PMST) and observed remaining survival time $TR$. In the PDF plots, vertical dashed black and blue lines represent the mode and median, respectively, of Weibull's PDF, which changes over time.

$S_t$: State of GRU-D cell at timestep t
$h_t$: hidden output at timestep t
$SP$: SoftPlus activation
$\tau$: Time remaining to event (or censoring)
$p(\tau)$: Probability of reaching end point at time $\tau$
$\kappa$: Shape parameter
$\lambda$: Scale parameter
$x$: Measurement value
$m$: Missing indicator
$d$: Delta time

a)

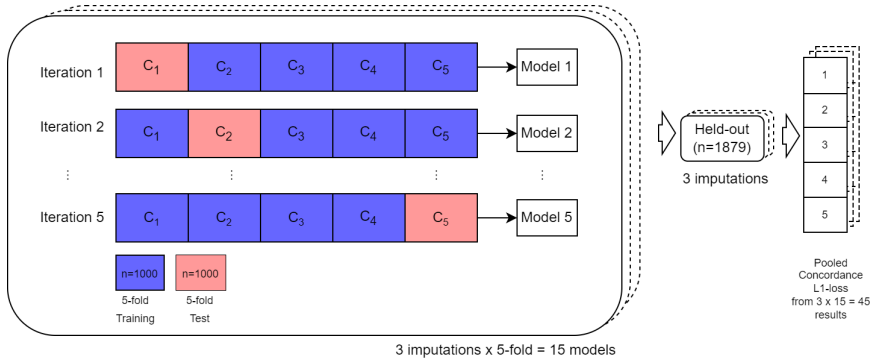

b)

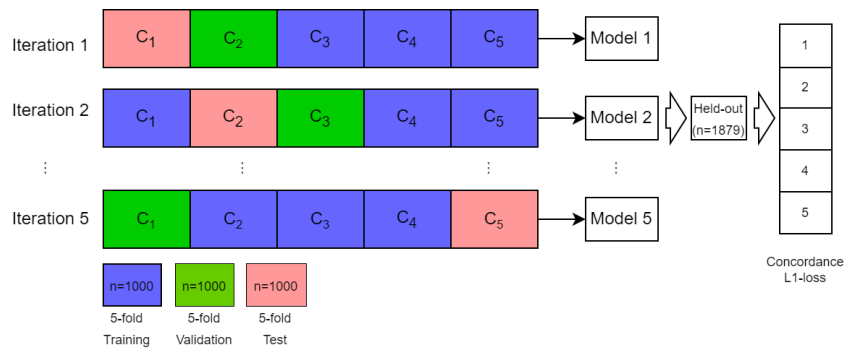

**Sup Fig 4** 5-fold cross validation and held-out strategy for a) AFT, XGB(AFT), RSF, and b) GRU-D-Weibull, GRU-lvcf-Weibull models. Specifically, XGB(AFT) uses an imputation scheme of a) and training scheme of b) since validation dataset is required for early stopping.

**References**

1. Chen, T. K., Knicely, D. H. & Grams, M. E. Chronic Kidney Disease Diagnosis and Management: A Review. *JAMA* **322**, 1294–1304 (2019).

2. Coresh, J. *et al.* Prevalence of chronic kidney disease in the United States. *JAMA* **298**, 2038–2047 (2007).

3. Plantinga, L. C. *et al.* Patient awareness of chronic kidney disease: trends and predictors. *Arch. Intern. Med.* **168**, 2268–2275 (2008).

4. Jha, V. *et al.* Chronic kidney disease: global dimension and perspectives. *Lancet* **382**, 260–272 (2013).

5. Hill, N. R. *et al.* Global Prevalence of Chronic Kidney Disease - A Systematic Review and Meta-Analysis. *PLoS One* **11**, e0158765 (2016).

6. Xie, Y. *et al.* Analysis of the Global Burden of Disease study highlights the global, regional, and national trends of chronic kidney disease epidemiology from 1990 to 2016. *Kidney Int.* **94**, 567–581 (2018).

7. Mills, K. T. *et al.* A systematic analysis of worldwide population-based data on the global burden of chronic kidney disease in 2010. *Kidney Int.* **88**, 950–957 (2015).

8. GBD 2015 Disease and Injury Incidence and Prevalence Collaborators. Global, regional, and national incidence, prevalence, and years lived with disability for 310 diseases and injuries, 1990-2015: a systematic analysis for the Global Burden of Disease Study 2015. *The Lancet* vol. 388 1545–1602 (2016).

9. Coresh, J. *et al.* Decline in estimated glomerular filtration rate and subsequent risk of end-stage renal disease and mortality. *JAMA* **311**, 2518–2531 (2014).

10. Johansen, K. L. *et al.* US Renal Data System 2020 Annual Data Report: Epidemiology of Kidney Disease in the United States. *Am. J. Kidney Dis.* **77**, A7–A8

(2021).

11. GBD 2017 Disease and Injury Incidence and Prevalence Collaborators. Global, regional, and national incidence, prevalence, and years lived with disability for 354 diseases and injuries for 195 countries and territories, 1990-2017: a systematic analysis for the Global Burden of Disease Study 2017. *Lancet* **392**, 1789–1858 (2018).

12. Kusiak, A., Dixon, B. & Shah, S. Predicting survival time for kidney dialysis patients: a data mining approach. *Comput. Biol. Med.* **35**, 311–327 (2005).

13. Tangri, N. A Predictive Model for Progression of Chronic Kidney Disease to Kidney Failure. *JAMA* vol. 305 1553 Preprint at https://doi.org/10.1001/jama.2011.451 (2011).

14. Noia, T. D. *et al.* An end stage kidney disease predictor based on an artificial neural networks ensemble. *Expert Systems with Applications* vol. 40 4438–4445 Preprint at https://doi.org/10.1016/j.eswa.2013.01.046 (2013).

15. Zhang, H., Hung, C.-L., Chu, W. C.-C., Chiu, P.-F. & Tang, C. Y. Chronic Kidney Disease Survival Prediction with Artificial Neural Networks. *2018 IEEE International Conference on Bioinformatics and Biomedicine (BIBM)* Preprint at https://doi.org/10.1109/bibm.2018.8621294 (2018).

16. Xiao, J. *et al.* Comparison and development of machine learning tools in the prediction of chronic kidney disease progression. *J. Transl. Med.* **17**, 119 (2019).

17. Naqvi, S. A. A., Tennankore, K., Vinson, A., Roy, P. C. & Abidi, S. S. R. Predicting Kidney Graft Survival Using Machine Learning Methods: Prediction Model Development and Feature Significance Analysis Study. *J. Med. Internet Res.* **23**, e26843 (2021).

18. Zou, Y. *et al.* Development and internal validation of machine learning algorithms for end-stage renal disease risk prediction model of people with type 2 diabetes mellitus and diabetic kidney disease. *Ren. Fail.* **44**, 562–570 (2022).

19. Lee, K.-H. *et al.* Artificial Intelligence for Risk Prediction of End-Stage Renal Disease in Sepsis Survivors with Chronic Kidney Disease. *Biomedicines* **10**, (2022).

20. Yuan, Q. *et al.* Role of Artificial Intelligence in Kidney Disease. *International Journal of Medical Sciences* vol. 17 970–984 Preprint at https://doi.org/10.7150/ijms.42078 (2020).

21. Katzman, J. L. *et al.* DeepSurv: personalized treatment recommender system using a Cox proportional hazards deep neural network. *BMC Med. Res. Methodol.* **18**, 24 (2018).

22. Ching, T., Zhu, X. & Garmire, L. X. Cox-nnet: An artificial neural network method for prognosis prediction of high-throughput omics data. *PLoS Comput. Biol.* **14**, e1006076 (2018).

23. Gensheimer, M. F. & Narasimhan, B. A scalable discrete-time survival model for neural networks. *PeerJ* **7**, e6257 (2019).

24. Che, Z., Purushotham, S., Cho, K., Sontag, D. & Liu, Y. Recurrent Neural Networks for Multivariate Time Series with Missing Values. *Sci. Rep.* **8**, 6085 (2018).

25. Chen, D. *et al.* Early Detection of Post-Surgical Complications using Time-series Electronic Health Records. *AMIA Jt Summits Transl Sci Proc* **2021**, 152–160 (2021).

26. Ruan, X. *et al.* Real-time risk prediction of colorectal surgery-related post-surgical complications using GRU-D model. *J. Biomed. Inform.* **135**, 104202 (2022).

27. Ruan, X. *et al.* Discrimination, calibration, and point estimate accuracy of

GRU-D-Weibull architecture for real-time individualized endpoint prediction. (2022) doi:10.48550/arXiv.2212.09606.

28. Ishwaran, H., Kogalur, U. B., Blackstone, E. H. & Lauer, M. S. Random survival forests. *Ann. Appl. Stat.* **2**, 841–860 (2008).

29. Tangri, N. *et al.* Multinational Assessment of Accuracy of Equations for Predicting Risk of Kidney Failure: A Meta-analysis. *JAMA* **315**, 164–174 (2016).

30. Levey, A. S. *et al.* A new equation to estimate glomerular filtration rate. *Ann. Intern. Med.* **150**, 604–612 (2009).

31. Healthcare Cost and Utilization Project (HCUP). *Encyclopedia of Health Services Research* Preprint at https://doi.org/10.4135/9781412971942.n164 (2009).

32. Henderson, R., Keiding, N., Københavns Universitet. Afdeling for Biostatistik & Afdeling, K. U. B. *Individual Survival Time Prediction Using Statistical Models*. (2004).

33. *Strictly Proper Scoring Rules, Prediction, and Estimation*. (2005).

34. Website. https://doi.org/10.48550/arXiv.1811.11347 doi:10.48550/arXiv.1811.11347.

35. Pickett, K. L., Suresh, K., Campbell, K. R., Davis, S. & Juarez-Colunga, E. Random survival forests for dynamic predictions of a time-to-event outcome using a longitudinal biomarker. *BMC Med. Res. Methodol.* **21**, 216 (2021).

36. Li, Q. & Xu, Y. VS-GRU: A Variable Sensitive Gated Recurrent Neural Network for Multivariate Time Series with Massive Missing Values. *Applied Sciences* vol. 9 3041 Preprint at https://doi.org/10.3390/app9153041 (2019).

37. Tsang, J. Y., Blakeman, T., Hegarty, J., Humphreys, J. & Harvey, G. Understanding the implementation of interventions to improve the management of chronic kidney disease in primary care: a rapid realist review. *Implement. Sci.* **11**, 47 (2016).

38. Barnwal, A., Cho, H. & Hocking, T. Survival regression with accelerated failure time model in XGBoost. *Journal of Computational and Graphical Statistics* 1–25 Preprint at https://doi.org/10.1080/10618600.2022.2067548 (2022).

39. Wang, D.-B., Feng, L. & Zhang, M.-L. Rethinking Calibration of Deep Neural Networks: Do Not Be Afraid of Overconfidence. *Adv. Neural Inf. Process. Syst.* **34**, 11809–11820 (2021).

40. Guo, C., Pleiss, G., Sun, Y. & Weinberger, K. Q. On Calibration of Modern Neural Networks. (2017).